\documentclass[letterpaper, 10 pt, conference]{ieeeconf}  % Comment this line out if you need a4paper

\IEEEoverridecommandlockouts                              % This command is only needed if 
\usepackage[colorlinks,urlcolor=blue,linkcolor=blue,citecolor=blue]{hyperref}
\usepackage{hyperref}

\title{\LARGE \bf HAViT: Historical Attention Vision Transformer}

\author{Swarnendu Banik$^{\ast}$, Manish Das$^{\S}$, Shiv Ram Dubey$^{\dagger}$, Satish Kumar Singh$^{\ddagger}$\\
Computer Vision and Biometrics Lab, Indian Institute of Information Technology, Allahabad\\
\{\tt\small $^{\ast}$mhc2023012, $^{\S}$pmm2023001, $^{\dagger}$srdubey, $^{\ddagger}$sk.singh\}@iiita.ac.in
}

\usepackage{graphicx}
\usepackage{algorithm}
\usepackage{algpseudocode}
\usepackage{subfig}
\usepackage{amssymb}

\usepackage{amsmath}
\usepackage{cleveref}
\usepackage{booktabs}

\usepackage{fancyhdr}

\fancypagestyle{firstpage}{%
  \rhead{2026 IEEE Conference on Artificial Intelligence (CAI)}
  \lfoot{\small{This paper is accepted by 2026 IEEE Conference on Artificial Intelligence (CAI). Copyright ©2026 IEEE. Personal use of this material is permitted. However, permission to use this material for any other purposes must be obtained from the IEEE by sending an email to pubs-permissions@ieee.org.}}
}

\begin{document}

\maketitle
\thispagestyle{firstpage}

%%%%%%%%%%%%%%%%%%%%%%%%%%%%%%%%%%%%%%%%%%%%%%%%%%%%%%%%%%%%%%%%%%%%%%%%%%%%%%%%
\begin{abstract}
Vision Transformers have excelled in computer vision but their attention mechanisms operate independently across layers, limiting information flow and feature learning. We propose an effective cross-layer attention propagation method that preserves and integrates historical attention matrices across encoder layers, offering a principled refinement of inter-layer information flow in Vision Transformers. This approach enables progressive refinement of attention patterns throughout the transformer hierarchy, enhancing feature acquisition and optimization dynamics. The method requires minimal architectural changes, adding only attention matrix storage and blending operations. Comprehensive experiments on CIFAR-100 and TinyImageNet demonstrate consistent accuracy improvements, with ViT performance increasing from 75.74\% to 77.07\% on CIFAR-100 (+1.33\%) and from 57.82\% to 59.07\% on TinyImageNet (+1.25\%). Cross-architecture validation shows similar gains across transformer variants, with CaiT showing 1.01\% enhancement. Systematic analysis identifies the blending hyperparameter of historical attention (\(\alpha = 0.45\)) as optimal across all configurations, providing the ideal balance between current and historical attention information. Random initialization consistently outperforms zero initialization, indicating that diverse initial attention patterns accelerate convergence and improve final performance.
Our code is publicly available at \url{https://github.com/banik-s/HAViT}
\end{abstract}

%%%%%%%%%%%%%%%%%%%%%%%%%%%%%%%%%%%%%%%%%%%%%%%%%%%%%%%%%%%%%%%%%%%%%%%%%%%%%%%%
\section{INTRODUCTION}

Vision Transformers (ViTs) \cite{dosovitskiy2020image} have fundamentally reshaped computer vision by showing that transformer architectures, originally designed for NLP, excel at image recognition and consistently rival or outperform traditional convolutional neural networks across tasks like classification, detection, and segmentation. Their success stems from self-attention \cite{dosovitskiy2020image}, which enables modeling of long-range spatial relationships between image patches for rich, contextual representation learning. This innovation has inspired diverse variants, such as Class-Attention Image Transformer (CaiT) \cite{touvron2021going} with class-attention, broadening the model’s capabilities.

Existing Vision Transformer implementations isolate attention layers, discarding valuable patterns post-computation and limiting hierarchical refinement. To address this gap, we introduce an innovative historical attention propagation methodology that preserves and integrates self-attention matrices (\(Q \!\cdot\! K^T\)) across sequential encoder layers. Our framework establishes an attention memory architecture that accumulates knowledge throughout network depth, offering a novel complement to existing enhancements. We validate this approach across diverse architectures (ViT \cite{dosovitskiy2020image}, CaiT \cite{touvron2021going}), demonstrating consistent gains on established benchmarks. On CIFAR-100, our method elevates baseline ViT accuracy from 75.74\% to 77.07\% (\(\uparrow\)1.33\%) with optimal \(\alpha = 0.45\) [Table~\ref{tab:cifar-100_alpha}]. Comparable improvements appear on TinyImageNet, rising from 57.82\% to 59.07\% (\(\uparrow\)1.25\%) [Table~\ref{tab:tinyimagenet_alpha}]. The approach also enhances CaiT performance from 73.85\% to 74.86\% (+1.01\%) [Table~\ref{tab:cross_arch}], confirming broad architectural compatibility.

\section{\textbf{Related Work}}

Vision Transformers (ViTs) have transformed computer vision by adapting NLP-based attention for visual tasks, with recent research focusing on architectural enhancements. Patro \emph{et al.} introduced SpectFormer, which uses spectral decomposition to capture global and local features \cite{patro2025spectformer}, while other works emphasize information flow beyond spatial dimensions, such as Channel Propagation Networks \cite{go2025channel} and localization masks in PatchGuard \cite{nafez2025patchguard}. Further innovations include GrafT \cite{Park_2024_WACV}, a lightweight add-on for multi-scale dependencies, and DeepViT's \cite{zhou2021deepvit} Re-attention, which regenerates attention maps to prevent collapse. Efficiency-focused designs like MicroViT \cite{setyawan2025microvit} and edge-optimized transformers \cite{saha2025vision} reduce complexity for constrained devices. Additionally, Cross-Layer Attention (CLA) \cite{brandon2405reducing} optimizes KV caching by sharing heads across layers, while HDPNet employs an hourglass dual-path architecture to refine global and spatial details for camouflaged object detection \cite{He_2025_WACV}. The big.LITTLE ViT \cite{guo2024big} introduces a dual-transformer architecture employing a dynamic inference mechanism to optimize efficiency. Dang \emph{et al.} utilize cascaded channel- and spatial-split transformers for facial landmark geometry modeling \cite{dang2025cascaded}. Yang \emph{et al.} propose focal self-attention \cite{yang2107focal} to capture both local and global interactions. The TORE framework \cite{olszewski2025tore} employs token recycling for active exploration, while SHViT \cite{yun2024shvit} utilizes a larger-stride patchify stem to reduce memory costs and enhance performance.

Recent work shows how similarity-guided adaptive transformers \cite{xue2025similarity} dynamically disable redundant layers for real-time UAV tracking. Further efficiency is achieved via softmax-free mechanisms \cite{koohpayegani2024sima}, focused linear attention \cite{han2023flatten}, and Less Attention ViT (LaViT) \cite{zhang2024you}, which minimizes attention operations. Ehab \emph{et al.} leverage feature fusion across DaViT, iFormer, and GPViT for deepfake detection \cite{essa2024feature}. Touvron \emph{et al.} \cite{touvron2021training} achieved data-efficient ViT training via knowledge distillation. Hierarchical architectures like Swin Transformer \cite{liu2021swin} utilize shifted windows for local attention, while FDViT \cite{xu2023fdvit} introduces flexible downsampling. ResNets \cite{He_2016_CVPR} established identity skip connections for deep training, a concept foundational to modern backbones. TransNeXt \cite{shi2024transnext} proposes aggregated biomimetic attention. Pre-training on small datasets is optimized by minimally scaled images \cite{tan2024pre,zhang2025depth}. 
A horizontally scalable vision transformer (HSViT) is introduced in \cite{xu2024hsvit} to tackle the collaborative model training and inference. SATA-ViT \cite{chen2023accumulated} performs the suppressing of accumulated trivial attention which is computed using attention importance.

Compact Convolutional Transformers (CCT) \cite{hassani2021escaping} and Patch Attention Excitation (PAE) \cite{verma2025patch} address small-dataset limitations via convolutional tokenization and channel-wise attention, respectively. 
Unlike DeepViT \cite{zhou2021deepvit}, which uses `Re-attention' to mix attention heads within the same layer (intra-layer) to boost diversity, HAViT explicitly propagates raw attention logits from the previous layer (inter-layer). DeepViT regenerates local patterns, whereas HAViT creates a direct memory path to preserve and refine historical context across the network depth, whereas HAViT blends historical attention matrices to progressively refine feature representations, serving a representational rather than computational goal.
Recent surveys \cite{han2022survey,nauen2308transformer} benchmark these ViT advancements against CNNs.

Despite recent advances in Vision Transformer architectures, a critical gap persists in inter-layer information flow: existing methods isolate layers, discarding valuable attention patterns after each computation. While approaches like channel propagation and token recycling explore connectivity, they fail to systematically preserve self-attention history for refinement. We address this by introducing an inter-layer attention propagation mechanism that enables historical pattern refinement throughout network depth. Unlike methods focused on individual layer optimization or architectural changes, our approach builds an attention memory system to accumulate and integrate self-attention knowledge across consecutive layers, offering a novel complement to existing performance enhancements.

\begin{figure*}[t]
  \centering
  \includegraphics[width=\textwidth]{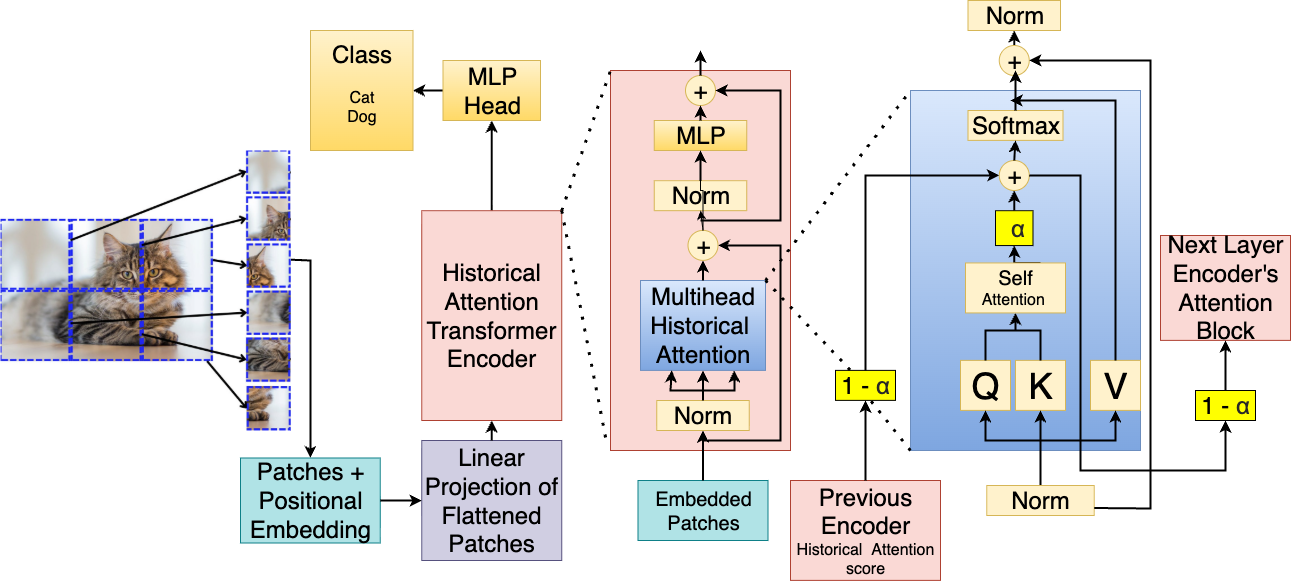}
  \caption{Overview of the modified ViT with historical attention propagation.}
  \label{fig:attention-propagation}
\end{figure*}

\section{\textbf{Proposed Historical Attention Mechanism}}

This section details our proposed cross-layer attention propagation framework for Vision Transformers. We commence by examining the conventional ViT architecture, subsequently presenting our innovative methodology for preserving and integrating unprocessed attention computations across sequential layers. Our approach resolves the core constraint of isolated attention processing in existing transformer designs by establishing attention memory that develops throughout network depth.

\subsection{\textbf{Conventional Vision Transformer Architecture}}

Vision Transformers handle images through decomposition into non-overlapping segments, treating individual segments as sequence tokens. For input images with dimensions $Height \times Width$ and segment size $P$, the total segment count equals:

\begin{equation}
N = \frac{Height \times Width}{P^2}
\end{equation}

Individual segments undergo linear transformation into $d_{model}$-dimensional representations, with a trainable classification token added to the sequence, resulting in a total sequence length of $n = N + 1$, for the CLS token (Classification token). The fundamental ViT component is the multi-head self-attention framework. Within each attention layer, the input sequence $X \in \mathbb{R}^{n \times d_{model}}$ generates query ($Q$), key ($K$), and value ($V$) representations:
\begin{equation}
Q = X W_Q,\quad
K = X W_K,\quad
V = X W_V
\end{equation}

where $W_{Q}, W_{K}, W_{V} \in \mathbb{R}^{d_{model} \times d_k}$ represent trainable transformation matrices, and $d_k = \frac{d_{model}}{h}$ denotes the per-head dimensionality with $h$ being the number of attention heads. The attention mechanism proceeds as:
\begin{equation}
\mathrm{Attention}(Q,K,V)
= \mathrm{softmax}\!\Bigl(\frac{QK^{T}}{\sqrt{d_k}}\Bigr)\,V
\end{equation}

This encodes similarity relationships among token pairs prior to softmax normalization. Standard ViT implementations immediately apply softmax to these values and subsequently eliminate them, precluding cross-layer attention pattern communication.

\subsection{\textbf{Historical Attention Propagation Architecture}}

As illustrated in Figure~\ref{fig:attention-propagation}, our methodology establishes a framework for maintaining and transmitting historical attention computations across sequential transformer layers. The fundamental principle recognizes that attention patterns from preceding layers encode valuable token relationship information that can enhance subsequent layer attention computations.

\paragraph{Mathematical Framework}
For attention layer $l$, we preserve the self attention matrix
\begin{equation}\label{eq:self_attention_scores}
A_{\text{self}}^{l} = \frac{Q^{l}(K^{l})^{T}}{\sqrt{d_k}}
\end{equation}

and transmit it to layer $l+1$. The attention computation in layer $l+1$ transforms to:
\begin{equation}\label{eq:attention_propagation}
A_{Current}^{\mathrm{l+1}}
= \alpha \cdot A_{Self}^{\mathrm{l+1}}
+ (1 - \alpha) \cdot H_{prev}^{\mathrm{l}}
\end{equation}

where \(H^{l}\) denotes the attention history from the preceding layer, i.e., 
\begin{equation}
H_{prev}^{\mathrm{l}} = A_{Current}^{\mathrm{l}}
\end{equation}

and \(\alpha\) is a fixed integration hyperparameter that controls the balance between current and historical attention information. For the first encoder layer, \(H_{\mathrm{0}}\) is initialized with random numbers (or zero initialization).

The final attention result computes as:
\begin{equation}
\mathrm{Attention}^{\mathrm{l+1}}
= \mathrm{softmax}\!\bigl(A_{Current}^{\mathrm{l+1}}\bigr)\,V^{\mathrm{l+1}}
\end{equation}
This formulation establishes attention continuity dynamics, enabling each layer to build upon previously established attention strategies while adapting to evolving feature representations. In the given algorithm (see Algorithm~\ref{alg:historical-attention}), it is shown how to compute the encoder output by blending current self‐attention with historical attention.

\begin{algorithm}[!t]
  \caption{Self-Attention with Historical Blending}
  \label{alg:historical-attention}
  \begin{algorithmic}[1]
    \Require Embeddings $X \in \mathbb{R}^{n \times d_{\text{model}}}$ 
    \Ensure Output and updated historical attention $H_{\text{current}}$
    \State Initialize blending coefficient  
        \Statex\quad $\displaystyle \alpha \;\leftarrow\;\mathcal{R}\in\,[0,1]$

    \If{this is the first encoder layer}
      \State $H_{\text{0}} \gets \text{Initialize}(\text{random or zero})$
    \Else
      \State $H_{\text{prev}} \gets H_{\text{previous layer}}$
    \EndIf
    \State $X \gets \mathrm{LayerNorm}(X)$
    \State $Q \gets X W_Q,\quad K \gets X W_K,\quad V \gets X W_V$
    \State $A_{\text{self}} \gets \dfrac{Q K^\top}{\sqrt{d_k}}$
    
    \State $A_{\text{current}} \gets \alpha \cdot A_{\text{self}} \;+\; (1 - \alpha)\cdot H_{\text{prev}}$
    \State $\mathrm{output} \gets \mathrm{softmax}\bigl(A_{\mathrm{Current}}\bigr)\,V$
    \State $H_{\text{current}} \gets A_{\text{current}}$
    \State \Return $\mathrm{output}$
  \end{algorithmic}
\end{algorithm}

\subsection{\textbf{Attention History Initialization Approaches}}

A fundamental design consideration involves initializing attention history $H_{0}$ for the initial transformer layer, where no preceding attention information exists. We examine two distinct initialization methodologies, each possessing different theoretical foundations and practical consequences.

\subsubsection{\textbf{Random Matrix Initialization}}

We initialize the attention history $H_{0}$ by sampling from a standard normal distribution, ensuring reproducibility via fixed seeds and separate random states for each run. This approach populates $H_{0}$ as:
\[
H_{0} \sim \mathcal{N}(0, I)
\;\in\;
\mathbb{R}^{B \times h \times n \times n}
\]
where $B$ is batch size, $h$ is head count, and $n$ denotes the total sequence length (patches + classification token).

\subsubsection{\textbf{Zero Matrix Initialization}}

We also initialize $H_{0}$ as a zero matrix, enabling the first layer to operate identically to a standard ViT initially. This strategy allows attention propagation effects to emerge gradually during training, helping to isolate the developmental benefits of accumulated attention history across network depth: $H_{0} = \mathbf{0}
\;\in\;
\mathbb{R}^{B \times h \times n \times n}
$.

\section{Experimental Setup}

This section presents the comprehensive experimental framework designed to evaluate our proposed historical attention propagation mechanism. The experimental design encompasses multiple architectural configurations, initialization strategies, and evaluation benchmarks to provide thorough assessment of our method's effectiveness.

\subsection{\textbf{Dataset Configuration and Preprocessing}}

Our evaluation employs two complementary benchmarks. CIFAR-100 \cite{krizhevsky2010cifar} provides 60,000 $32\times32$ color images across 100 classes, split into 50,000 training and 10,000 test samples. TinyImageNet \cite{le2015tiny} offers 200 classes, containing 100,000 training and 10,000 validation images (500 and 50 per class, respectively), all resized to $64\times64$.

\subsection{Training Settings}
To ensure a fair comparison, we train both ViT and CaiT backbones using the same setup. Each model is trained for 100 epochs with AdamW (learning rate 0.003, weight decay $5\times10^{-2}$, batch size 128), along with a cosine learning rate schedule and a 10-epoch warm-up. We also apply label smoothing with a factor of 0.1. To keep the token count consistent across datasets, we adapt the patch size based on input resolution: 4$\times$4 patches for 32$\times$32 CIFAR-100 and 8$\times$8 patches for 64$\times$64 TinyImageNet. All optimization and regularization settings are kept identical so that any performance differences can be attributed only to architectural changes. We further study the effect of the blending parameter $\alpha$ by varying it from 0.1 to 0.9, and evaluate each setting using both random and zero initialization across all models and datasets.

\section{\textbf{Results and Analysis}}

This section presents experimental results demonstrating our historical attention propagation mechanism's effectiveness, analyzed across datasets, architectures, and initialization strategies for understanding performance characteristics and optimal configurations.

\subsection{\textbf{CIFAR-100 Performance Analysis}}

\begin{figure}[!t]
    \centering
    \includegraphics[width=0.488\columnwidth]{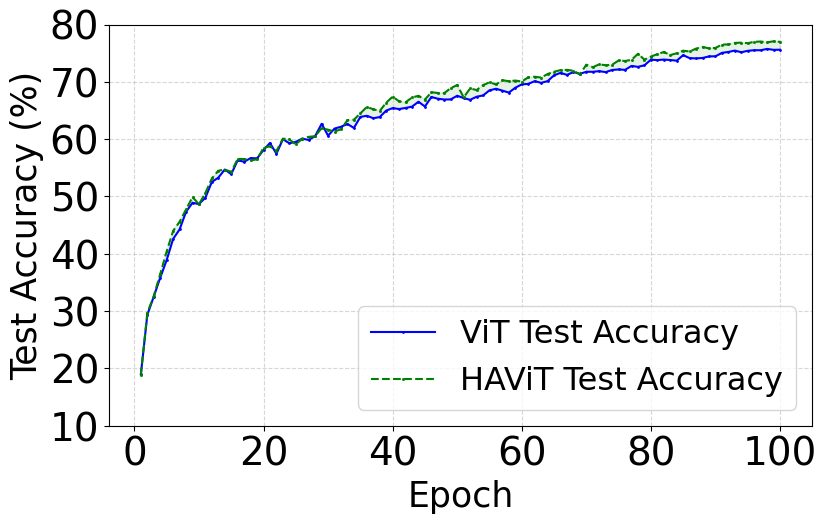}
    \includegraphics[width=0.488\columnwidth]{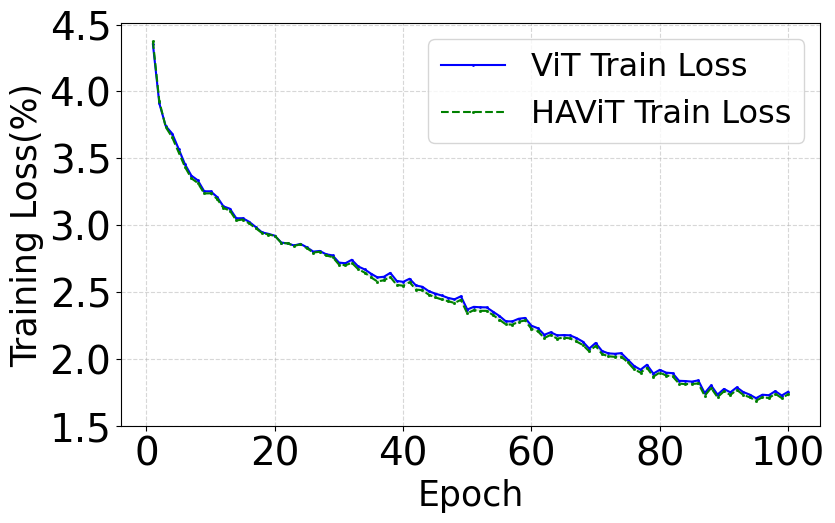}
    \caption{(a) Test accuracy comparison over epochs for CIFAR-100 dataset. 
    (b) Training loss comparison over epochs for CIFAR-100 dataset.}
    \label{fig:cifar-100-graph}
\end{figure}

Table~\ref{tab:cifar-100_alpha} evaluates HAViT's performance on CIFAR-100. Baseline ViT attains 75.74\% accuracy, while HAViT with random initialization peaks at 77.07\% (\(\uparrow\)1.33\%) at \(\alpha=0.45\), demonstrating effective inter-layer attention blending (see Fig.~\ref{fig:random-zero}(a)). Zero initialization also yields gains, reaching 76.50\% (\(\uparrow\)0.76\%), though the 0.57\% gap highlights the benefit of random initialization. Sensitivity analysis reveals an inverted-U trend where mid-range \(\alpha\) values (0.40–0.50) optimize performance, whereas extremes (\(<\)0.25 or \(>\)0.75) limit adaptability. Fig.~\ref{fig:cifar-100-graph} confirms HAViT's stable convergence and consistent improvement over the baseline.

\begin{table}[!t]
  \centering
  \caption{CIFAR-100 Results with $\alpha$ Variation (Baseline ViT accuracy = 75.74)}
  \normalsize
  \label{tab:cifar-100_alpha}
  \begin{tabular}{p{0.15\columnwidth} p{0.35\columnwidth} p{0.35\columnwidth}}
      \toprule
      \(\alpha\) & \textbf{Accuracy} & \textbf{Accuracy(\%)} \\
                 & \textbf{(Random)}   & \textbf{(Zero)} \\
      \midrule
      0.10 & 76.14 (\(\uparrow\)0.40) & 75.83 (\(\uparrow\)0.09) \\
      0.25 & 76.29 (\(\uparrow\)0.55) & 76.02 (\(\uparrow\)0.28) \\
      0.40 & 76.85 (\(\uparrow\)1.11) & 76.25 (\(\uparrow\)0.51) \\
      0.45 & 77.07 (\(\uparrow\)1.33) & 76.50 (\(\uparrow\)0.76) \\
      0.50 & 76.43 (\(\uparrow\)0.69) & 76.41 (\(\uparrow\)0.67) \\
      0.60 & 76.25 (\(\uparrow\)0.51) & 76.67 (\(\uparrow\)0.93) \\
      0.75 & 76.08 (\(\uparrow\)0.34) & 75.91 (\(\uparrow\)0.17) \\
      0.80 & 75.76 (\(\uparrow\)0.02) & 76.11 (\(\uparrow\)0.37) \\
      0.90 & 75.77 (\(\uparrow\)0.03) & 76.35 (\(\uparrow\)0.61) \\
      \bottomrule
  \end{tabular}
\end{table}

\subsection{\textbf{TinyImageNet Performance Evaluation}}

\begin{figure}[!t]
    \centering
    \includegraphics[width=0.489\columnwidth]{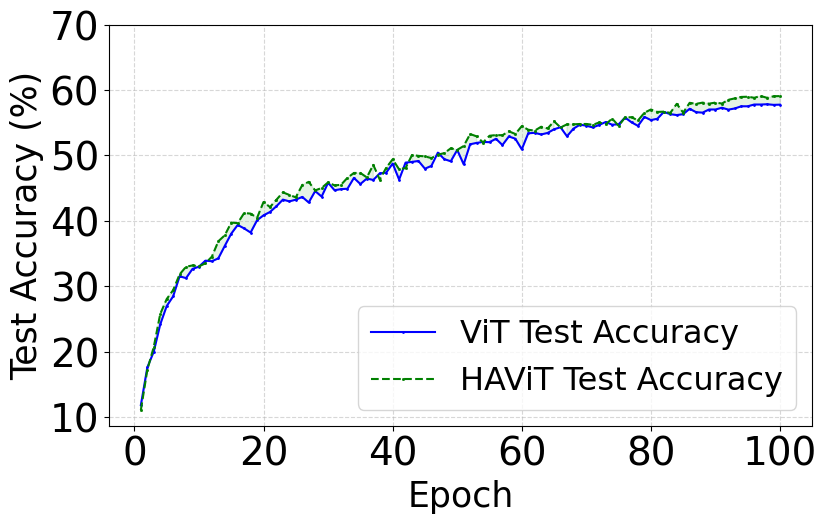}
    \includegraphics[width=0.489\columnwidth]{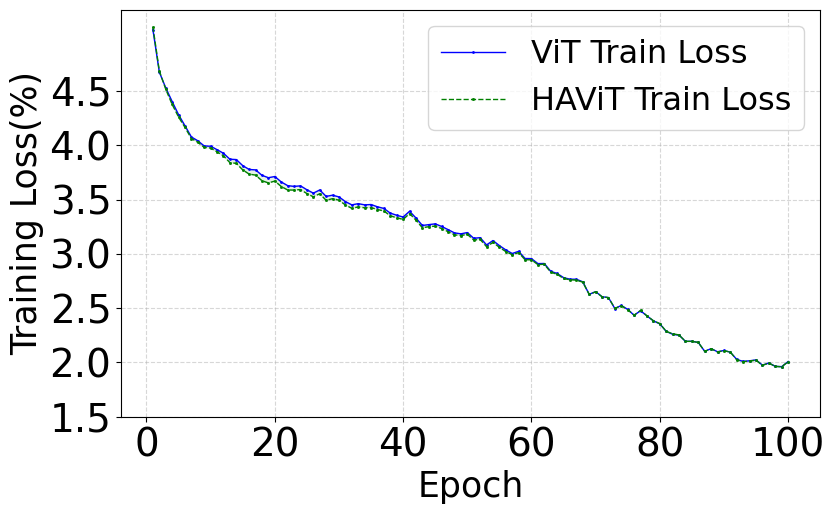}
    \caption{(a) Test accuracy comparison over epochs for TinyImageNet dataset.
    (b) Training loss comparison over epochs for TinyImageNet dataset.}
    \label{fig:tinyimagenet-graph}
\end{figure}

\begin{figure}[!t]
    \centering
    \includegraphics[width=0.485\columnwidth]{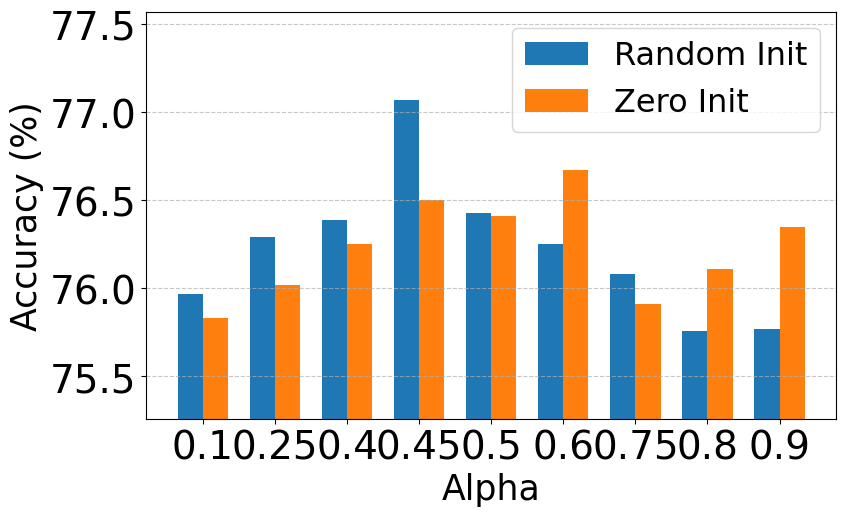}
    \includegraphics[width=0.485\columnwidth]{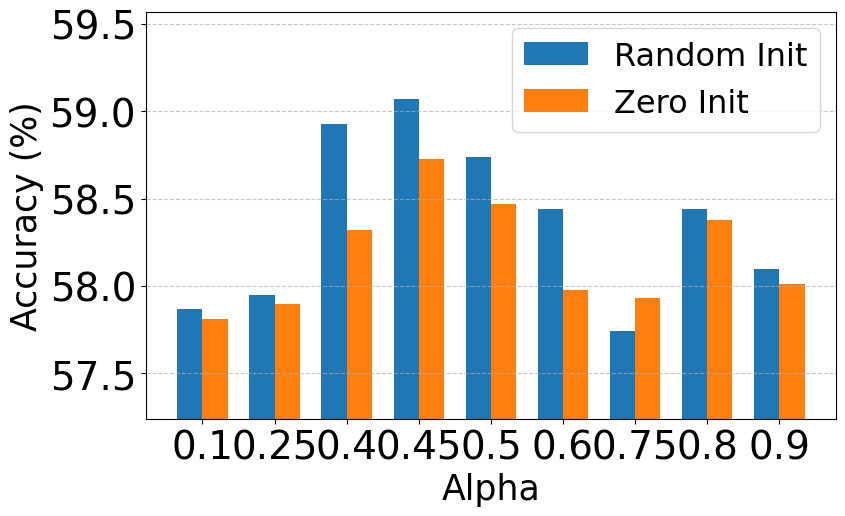}
    \caption{Initialization Strategy Impact Analysis:
             (a) Accuracy for $\alpha$ on CIFAR-100 (ViT baseline 75.74\%),
             (b) Accuracy for $\alpha$ on TinyImageNet (ViT baseline 57.82\%).}
    \label{fig:random-zero}
\end{figure}

Table~\ref{tab:tinyimagenet_alpha} confirms HAViT's scalability on TinyImageNet, achieving 59.07\% accuracy compared to the baseline ViT's 57.82\%. Performance analysis shows a strong dependency on the blending hyperparameter~\(\alpha\), peaking at \(\alpha=0.45\) with 59.07\% (\(\uparrow\)1.25\%) for random initialization and 58.73\% (\(\uparrow\)0.91\%) for zero initialization (see Fig.~\ref{fig:random-zero}(b)). This highlights that a balanced mix of current and historical attention optimizes feature learning. Furthermore, random initialization consistently outperforms zero initialization, notably widening the gap by 0.34\% at the optimal \(\alpha\), suggesting that controlled randomness aids the historical attention mechanism. Fig.~\ref{fig:tinyimagenet-graph} illustrates the stable convergence and superior training dynamics of HAViT over the baseline.

\begin{table}[!t]
  \centering
  \caption{TinyImageNet  Results with $\alpha$ Variation (Baseline ViT accuracy = 57.82)}
  \normalsize
  \label{tab:tinyimagenet_alpha}
  \begin{tabular}{p{0.15\columnwidth} p{0.35\columnwidth} p{0.35\columnwidth}}
      \toprule
      \(\alpha\) & \textbf{Accuracy} & \textbf{Accuracy (\%)} \\
                 & \textbf{(Random)}   & \textbf{(Zero)} \\
      \midrule
    0.10 & 57.79 (↓0.03) & 57.89 (↑0.07) \\
    0.25 & 58.21 (↑0.39) & 57.90 (↑0.08) \\
    0.40 & 58.93 (↑1.11) & 58.32 (↑0.50) \\
    0.45 & 59.07 (↑1.25) & 58.73 (↑0.91) \\
    0.50 & 58.74 (↑0.92) & 58.47 (↑0.65) \\
    0.60 & 58.44 (↑0.62) & 57.98 (↑0.16) \\
    0.75 & 57.74 (↓0.08) & 57.93 (↑0.11) \\
    0.80 & 58.44 (↑0.62) & 58.38 (↑0.56) \\
    0.90 & 58.01 (↑0.19) & 58.01 (↑0.19) \\
    \bottomrule
  \end{tabular}
\end{table}

\begin{figure*}[!t]
  \centering
  \includegraphics[width=\textwidth]{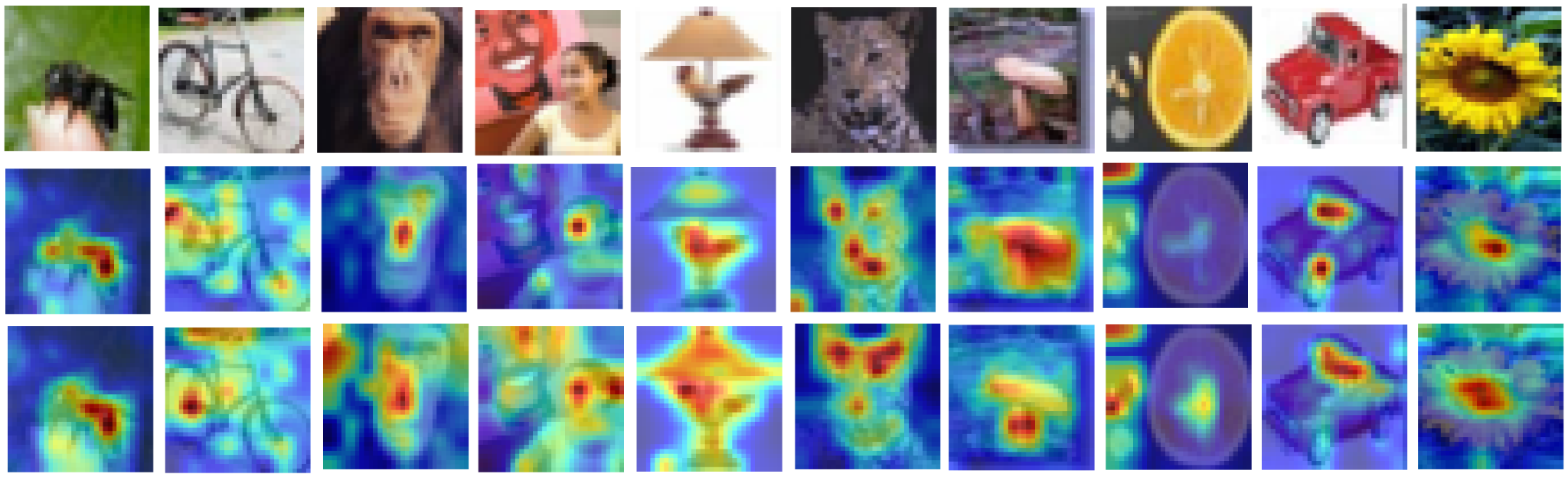}
  \caption{Attention map comparison between Base ViT and HAViT across CIFAR-100 classes}
  \label{fig:Attention viz}
\end{figure*}

\subsection{\textbf{Cross-Architecture Performance Validation}}

Table~\ref{tab:cross_arch} compares performance across multiple Vision Transformer variants using the optimal setup ($\alpha=0.45$, random initialization). All architectures (ViT \cite{dosovitskiy2020image}, CaiT \cite{touvron2021going}) benefit from historical attention propagation, with improvements from 1.01\% to 1.33\%, confirming the method’s broad applicability.

\begin{table}[!t]
  \centering
  \caption{{Cross-Architecture Comparison (optimal $\alpha=0.45$, random initialization.})}
  \label{tab:cross_arch}
  \resizebox{\columnwidth}{!}{%
  \begin{tabular}{@{}l l c c@{}}
    \toprule
    \textbf{Architecture} & \textbf{Dataset}      & \textbf{Baseline (\%)} & \textbf{Modified (\%)} \\
    \midrule
    ViT          & CIFAR-100    & 75.74 & 77.07 \\
    CaiT         & CIFAR-100    & 73.85 & 74.86 \\
    ViT          & TinyImageNet & 57.82 & 59.07 \\
    CaiT         & TinyImageNet & 58.01 & 59.02 \\

    \bottomrule
  \end{tabular}
  }
\end{table}

\subsection{\textbf{Results Comparison with State-of-the-Art Models on CIFAR-100 Dataset}}

Table~\ref{tab:comparison-existing} benchmarks HAViT on CIFAR-100, where it achieves 77.07\% accuracy, a strong result against established ViT and CNN baselines. It outperforms standard Vision Transformers like ViT-Lite-7/4 (73.94\%), Compact Vision Transformers like CVT-7/4 (76.49\%) and Compact Convolution ViT like CCT-2/3x2 (66.93\%) \cite{hassani2021escaping}, as well as classic CNNs like ResNet-56 (74.81\%) and ResNet-110 (76.63\%) \cite{He_2016_CVPR}. While method like CCT-7/3x2 (77.72\%)  \cite{hassani2021escaping} 
% and Mae-ViT-C100 (78.27\%) \cite{tan2024pre}
score slightly higher, it requires complex pretraining. In contrast, HAViT remains competitive solely through simple architectural changes, avoiding the need for elaborate pre-training schemes.

\begin{table}[!t]
  \centering
  \caption{Accuracy Comparison with SOTA Models on CIFAR-100 Dataset}
  \label{tab:comparison-existing}
  \resizebox{\columnwidth}{!}{%
  \begin{tabular}{@{}l l c@{}}
    \toprule
    \textbf{Method} & \textbf{Architecture Type} & \textbf{Accuracy (\%)} \\
    \midrule
    ResNet-56 \cite{He_2016_CVPR}                & CNN                          & 74.81 \\
    ResNet-110 \cite{He_2016_CVPR}                & CNN                          & 76.63 \\
    ViT-Lite-7/4 \cite{hassani2021escaping}              & Standard ViT                 & 73.94 \\
    CVT-7/4 \cite{hassani2021escaping}                   & Compact ViT                  & 76.49 \\
    CCT-2/3×2 \cite{hassani2021escaping}                 & Compact Convolutional ViT    & 66.93 \\
    CCT-7/3×2 \cite{hassani2021escaping}                 & Compact Convolutional ViT    & 77.72 \\
    % Mae-ViT-C100 \cite{tan2024pre}              & Pretrained ViT               & 78.27 \\
    \textbf{HAViT (Ours)}      & \textbf{Modified ViT }       & \textbf{77.07} \\
    \bottomrule
  \end{tabular}%
  }
\end{table}

\begin{table}[!t]
  \centering
  \caption{Accuracy Comparison with SOTA Models on TinyImageNet Dataset}
  \label{tab:tinyimagenet_full_comparison}
  \resizebox{1\columnwidth}{!}{%
    \begin{tabular}{@{}l l c@{}}
    \toprule
    \textbf{Method}                       & \textbf{Architecture Type}   & \textbf{Accuracy (\%)} \\
    \midrule
    ResNet-18 \cite{He_2016_CVPR}         & CNN                          & 53.32 \\
    ResNet-56 \cite{He_2016_CVPR}         & CNN                          & 58.77 \\
    % ResNet-110 \cite{He_2016_CVPR}        & CNN                          & 62.96 \\
    ViT (baseline) \cite{dosovitskiy2020image}    & Standard ViT & 57.82 \\
    % ViT-Tiny (baseline) \cite{touvron2021training}    & Standard ViT        & 59.00 \\
   HSViT-C3A4 \cite{xu2024hsvit}  & Modified ViT     & 56.73 \\
   SATA-ViT \cite{chen2023accumulated}  & Modified ViT     & 58.77 \\
   \textbf{HAViT (Ours)}                          & \textbf{Modified ViT }     & \textbf{59.07} \\
    % SL-ViT \cite{lee2021vision}                   & Enhanced ViT        & 61.07 \\
    % Swin (scratch) \dag{} \cite{liu2021swin}              & Hierarchical ViT    & 60.05  \\
    % ViT-DRLoc \cite{liu2021efficient}                & Standard ViT        & 42.33 \\
    \bottomrule
  \end{tabular}
  }
\end{table}

\subsection{\textbf{Results Comparison with State-of-the-Art Models on TinyImageNet Dataset}}

Table~\ref{tab:tinyimagenet_full_comparison} details our evaluation on TinyImageNet, where HAViT reaches 59.07\% accuracy. This positions it competitively against specialized architectures, especially considering HAViT relies on a simple architectural modification rather than complex training schemes. The proposed HAViT model is able to outperform the SOTA models, such as ResNet-18 (53.22\%), ResNet-56 (58.77\%), ViT (57.82\%), HSViT (56.73\%) and SATA-ViT (58.77\%). It shows that the proposed historical attention provides the useful past information while future transformer encoder layers process the embeddings.
% For instance, while SL-ViT \cite{lee2021vision} achieves 61.07\% via shifted patch tokenization and Gani \emph{et al.}\ \cite{gani2022train} reach 63.36\% using two-stage self-supervision, HAViT remains comparable without such overhead. It outperforms standard CNNs like ResNet-56 (58.77\%) \cite{He_2016_CVPR} and approaches deeper baselines such as ResNet-110 (62.96\%) and EfficientNet-B0 (66.79\%) \cite{tan2024pre}, demonstrating robustness on small datasets without additional inductive biases.

\subsection{\textbf{Attention Visualization Analysis}}

Fig.~\ref{fig:Attention viz} highlights distinct attention dynamics: while Base ViT tends to fixate on central object regions, HAViT produces a more distributed attention pattern due to inter-layer blending. This expanded coverage allows the model to capture critical context and background details that the baseline often misses. The difference is particularly sharp in complex scenes such as those with bicycles or multiple subjects where HAViT maintains awareness of both foreground and background elements, leading to richer feature representation.

\section{\textbf{Conclusion}}

We have presented a novel historical attention propagation mechanism for Vision Transformers that overcomes the limitation of independent attention computations by preserving and blending historical attention matrices across layers. Our approach introduces an \emph{attention memory} system, where each layer’s self attention \(A_{\mathrm{self}}\) is combined with the previous layer’s history \(H_{\mathrm{prev}}\) 
and we find \(\alpha = 0.45\) to be universally optimal across ViT, CaiT architectures on CIFAR-100 and TinyImageNet. Random initialization of \(H_0\) consistently outperforms zero initialization, accelerating convergence and boosting accuracy by up to 1.33\% over baseline models. Cross-architecture experiments confirm the broad applicability of our method, with CaiT gaining +1.01\%. These results demonstrate practical benefits with minimal architectural overhead and offer insights into attention dynamics, highlighting the value of inter-layer historical information sharing and contributing an effective refinement to transformer design by establishing attention continuity as a lightweight complement to existing deep learning architectures.

\section{ACKNOWLEDGMENT}

We gratefully acknowledge the Indian Institute of Information Technology Allahabad, Ministry of Education, Govt.of India, for providing the fellowship to pursue this work.

\bibliographystyle{IEEEtran}
\bibliography{ref}

\end{document}